\documentclass{article}
\usepackage{spconf,amsmath,graphicx}

\usepackage{graphicx}
\usepackage{amsmath,amssymb,amsfonts}
\usepackage{booktabs}
\usepackage{multirow,multicol}
\usepackage[table,xcdraw]{xcolor}
\usepackage{listings}
\usepackage{algorithm}
\usepackage{algorithmicx,algpseudocode}
\usepackage[caption=false]{subfig}
\usepackage{array}
\usepackage{bm}
\usepackage{tikz}
\usetikzlibrary{bayesnet,matrix}
\usepackage{longtable,rotating}
\usepackage{threeparttable}

\usepackage[normalem]{ulem}

\useunder{\uline}{\ul}{}

\title{Flattening Singular Values of Factorized Convolution for Medical Images}
\name{Zexin Feng$^{\rm 1}$\sthanks{The two authors contributed equally to this paper.}, Na Zeng$^{\rm 1}$\footnotemark[1], Jiansheng Fang$^{\rm 2}$, Xingyue Wang$^{\rm 1}$, Xiaoxi Lu$^{\rm 1}$, Heng Meng$^{\rm 3}$\sthanks{Corresponding author.}, Jiang Liu$^{\rm 1}$\footnotemark[2]}
\address{Author Affiliation(s)}
\address{$^{\rm 1}$Research Institute of Trustworthy Autonomous Systems and Department of Computer Science\\
and Engineering, Southern University of Science and Technology, Shenzhen, China\\
$^{\rm 2}$Guangzhou Native-Stone Intelligent-Brain Technology Co., Ltd., Guangzhou, China\\
$^{\rm 3}$Department of Neurology, The First Affiliated Hospital of Jinan University, Guangzhou, China}
\begin{document}

%
\maketitle
\begin{abstract}
Convolutional neural networks (CNNs) have long been the paradigm of choice for robust medical image processing (MIP). Therefore, it is crucial to effectively and efficiently deploy CNNs on devices with different computing capabilities to support computer-aided diagnosis. Many methods employ factorized convolutional layers to alleviate the burden of limited computational resources at the expense of expressiveness. To this end, given weak medical image-driven CNN model optimization, a Singular value equalization generalizer-induced Factorized Convolution (SFConv) is proposed to improve the expressive power of factorized convolutions in MIP models. We first decompose the weight matrix of convolutional filters into two low-rank matrices to achieve model reduction. Then minimize the KL divergence between the two low-rank weight matrices and the uniform distribution, thereby reducing the number of singular value directions with significant variance. Extensive experiments on fundus and OCTA datasets demonstrate that our SFConv yields competitive expressiveness over vanilla convolutions while reducing complexity.

\end{abstract}
\renewcommand{\thefootnote}{\fnsymbol{footnote}}
\footnotetext[3]{This work was supported in part by General Program of National Natural Science Foundation of China (Grant No. 82272086), and Shenzhen Natural Science Fund (JCYJ20200109140820699 and the Stable Support Plan Program 20200925174052004).}

\section{Introduction}
\label{sec:intro}

The representation power of convolutional neural networks (CNNs) delivers impressive prospects in computer-aided medical image diagnosis. Hence, in assisting clinical decision-making practicality, it is essential to effectively and efficiently deploy CNNs for medical image processing (MIP) on devices with different computing capabilities. However, the complexity of CNN models significantly limits their deployment on available devices, thus fading the progress of automatic MIP. To this end, some CNN model compression methods have been explored to reduce complexity, including network pruning~\cite{han2015learning}, knowledge distillation~\cite{hinton2015distilling}, and low-rank decomposition~\cite{zhang2015accelerating}. 
Low-rank decomposition is widely used in model compression for its simplicity, ease of implementation~\cite{wang2017factorized,li2019compressing}, and notable reduction in storage and computational costs~\cite{denton2014exploiting}.

\begin{figure}[t]
\centering
\includegraphics[width=0.95\linewidth]{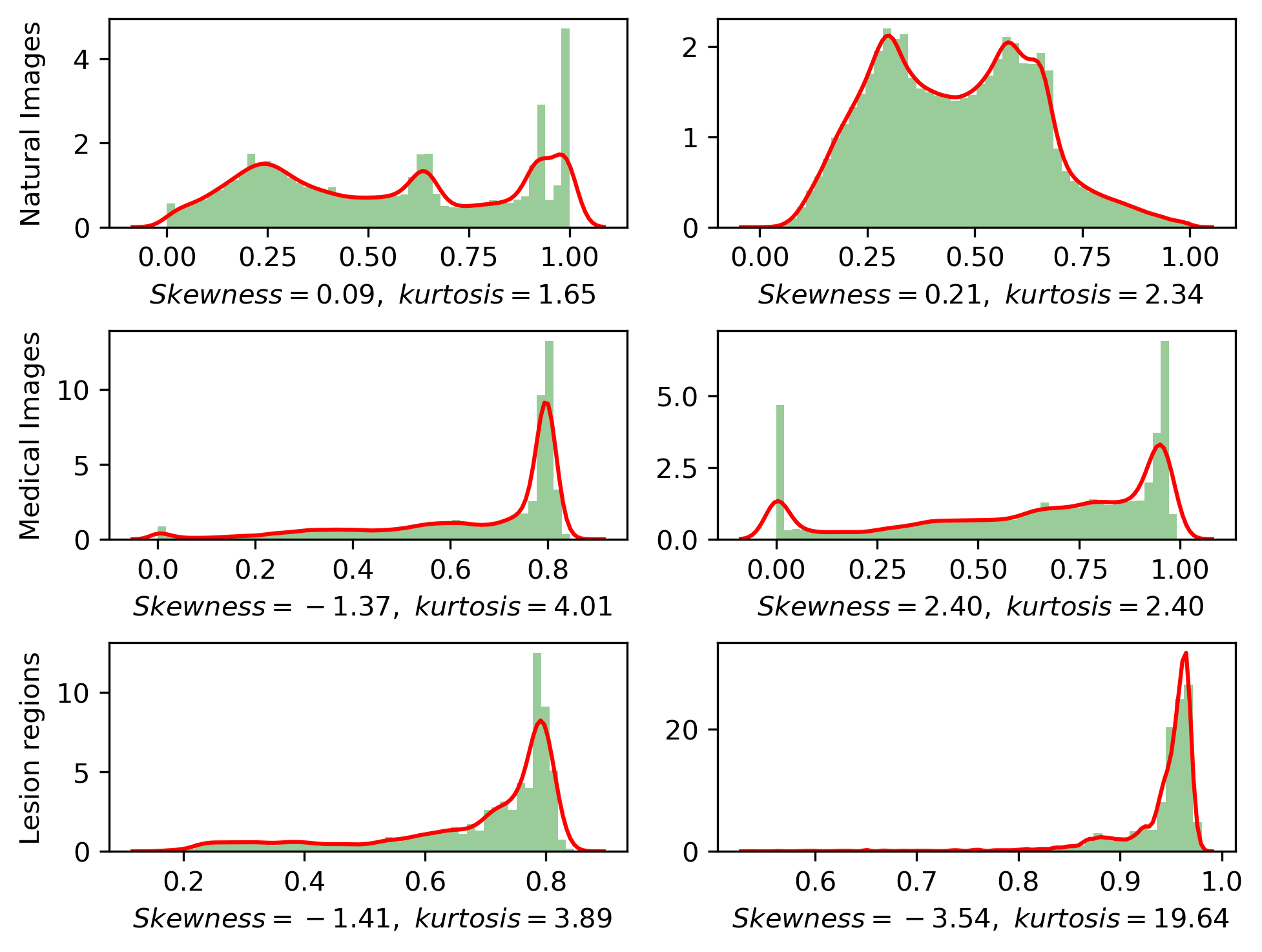} 
\caption{Measures of skewness and kurtosis of data distribution by calculating the histogram of pixel values. The first row is two natural images randomly selected from the VOC2012 dataset. The second row is two chest X-rays randomly sampled from the VIN-CXR dataset, and their corresponding lesion regions are in the third row.}
\label{fig:hist_vis}
\end{figure}

The intuition of low-rank decomposition is to reparametrize the weight of a convolution filter by the products of two low-dimensional matrices to alleviate the heavy burden of model complexity. Despite such an appealing property of matrix factorization, the model expressiveness declines for the significant compression of weight parameters to be optimized by the gradient. To this end, some work explores improving the model expressiveness of factorized convolutions for MIP. An inherited work~\cite{khodak2021initialization} focuses on the initialization and regularization of factorized layers and studies how they interact with gradient-based optimization of factorized layers. Another work~\cite{zeng2022factorized} exploits spectral normalization to optimize the expressive power of factorized convolution to meet the needs of model compression in medical image processing.
However, these works above have two limitations:
Firstly, they employ matrix factorization to compress the parameters of the convolution kernel, in which the speedup of inference time is bounded by the matrix multiplication computation. 
Secondly, these methods lack comprehensive exploration of the characteristics of medical images, so the optimization schemes they adopt are sub-optimal.
We count up the average skewnesses of pixel distribution for each image in 10,000 images randomly sampled from the VOC2012~\cite{everingham2015pascal} and VIN-CXR dataset~\cite{nguyen2020vindr}. 
Skewness is a measure of symmetry, and the skewness of symmetric distribution approximates zero~\cite{doane2011measuring}. As shown in Fig.\ref{fig:hist_vis}, the distributions of two medical images are heavy skewness and high kurtosis compare to the two natural images. 
Analyzing pixel patterns, we find that the spatial redundancy of medical images is heavier than natural images, implying that missing pixels can be reconstructed from neighbors. The low pixel-wise variance of medical images leads to weaker driving ability of the model than natural images.


To address above problems, we propose a novel Singular value equalization generalizer-induced Factorized Convolution (SFConv). With the purpose of avoiding extra computation from multiplying low-rank matrices, two low-rank convolutions is employed instead of one high-rank convolution for low-rank decomposition~\cite{tai2015convolutional}.

In addtion, inspired by the pixel distribution properties of medical images, a singular value equalization generalizer is introduced to enhance factorized convolutions in MIP models.
This regularizer addresses lower variance in medical images by flattening singular values of weight matrices in two low-rank convolutions through KL divergence-based penalty, improving weight variance and preventing the occurrence of overly large singular values.

The main contributions of this work are: (1) We leverage two low-rank convolutions to achieve low-rank decomposition, thereby avoiding the decline of inference speed caused by matrix multiplication calculation; (2) We design a KL regularizer based on the properties of medical images to improve the expressiveness of factorized convolutions by flattening singular values; (3) We conduct extensive experiments on fundus and OCTA datasets to demonstrate the effectiveness of our SFConv.

\section{Methodology}
\label{sec:met}
\subsection{Factorized Convolution}
\label{sec:fc}
Consider the input feature $\bm{I}_{in} \in \mathbb{R}^{c_{in}\times h \times w}$ and the convolutional filter $\bm{W} \in \mathbb{R}^{c_{out}\times c_{in} \times k \times k}$, where $h$ and $w$ are the height and width of the input feature, $c_{in}$ and $c_{out}$ denote the number of input and output channels, $k$ is the kernel size of the convolution. 
The reduction in model complexity is achieved by decomposing the parameters of the convolutional layers.
We align convolutional parameters of $c_{out}c_{in}k^{2}$ as a weight matrix of size $m \times n$, where $m = c_{out} \times k$ and $n = c_{in}\times k$. Low-rank decomposition is applied to project the weight matrix into two low-rank matrices with a joint latent space. Assuming the dimension of the joint latent space is $r$, and it varies in different tasks and meets the condition of $r \ll \min(m, n)$. The weight matrix $\bm{W}$ is factorized as:
\begin{equation}
\bm{W}_{mn}\approx \bm{P}_{mr} \times \bm{Q}_{rn}
\label{equ:wmn},
\end{equation}
where $\bm{P}$ and $\bm{Q}$ denote the two projected matrices with rank-$r$. The output feature $\bm{I}_{out}$ of feed-forward convolution is obtained by the function $\bm{I}_{out} = f(\bm{I}_{in}, \bm{W})$. After low-rank decomposition, the function is rewritten as:
\begin{equation}
\bm{I}_{out} = f(\bm{I}_{in}, \bm{P}, \bm{Q})
\label{equ:io1}.
\end{equation}

\begin{figure}[t]
\centering
\includegraphics[width=1.05\linewidth]{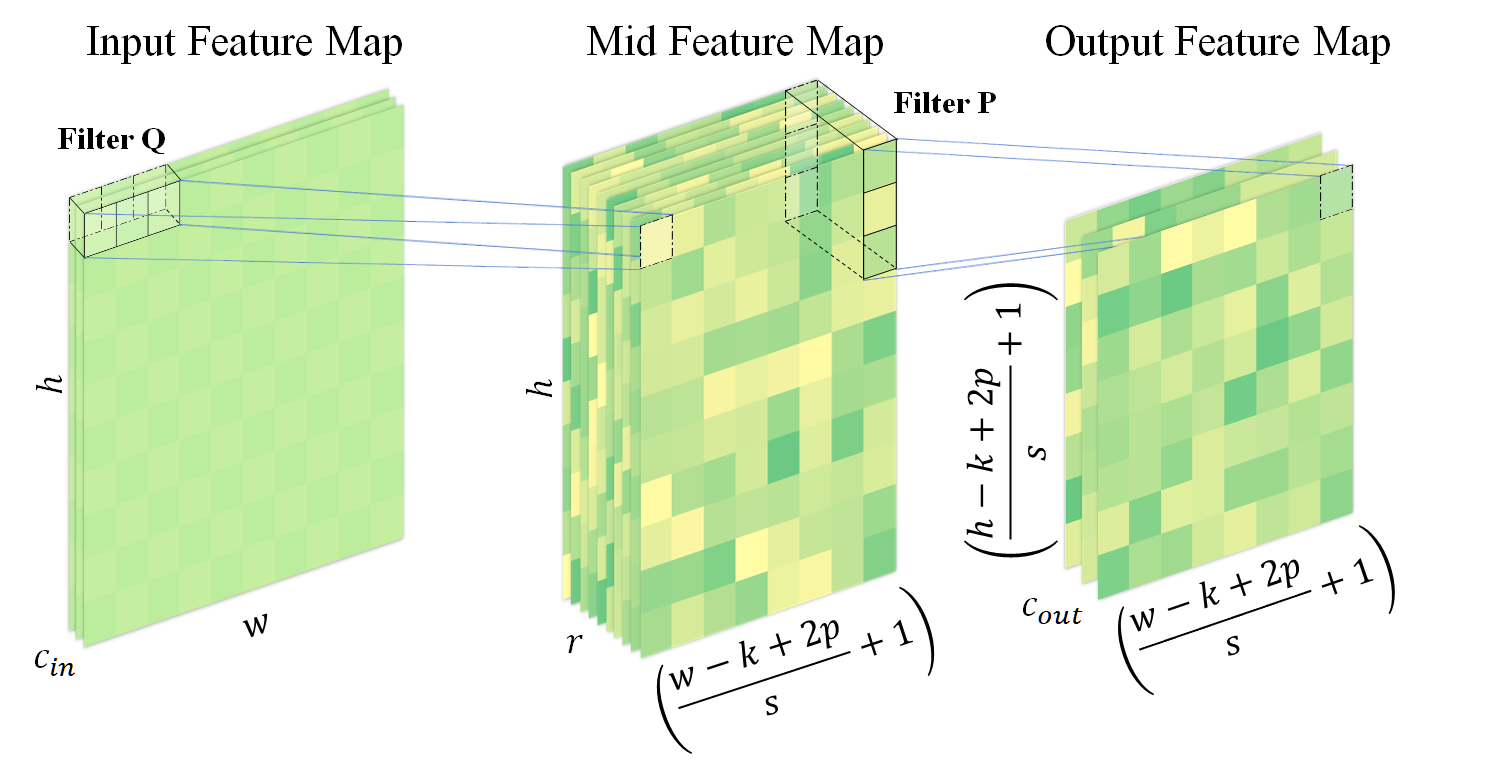} 
\caption{Schematic diagram of  SFConv, where $s$ is stride and $p$ is padding. The other variables are described in section~\ref{sec:fc}. }
\label{fig:SFConv}
\end{figure}

Although the number of parameters requiring gradient computation significantly decreases, the computation overhead corresponding increases due to the product of $\bm{P}$ and $\bm{Q}$. Hence, we adopt the composition of two 1d convolutions to implement matrix multiplication by reshaping $\bm{Q} \in \mathbb{R}^{r \times c_{in} \times 1 \times k} $ consisting of $r$ output channels and filters of size $1 \times k$, and $\bm{P} \in \mathbb{R}^{c_{out} \times r \times 1 \times k} $ consisting of $c_{out}$ output channels and filters of size $1 \times k$. As shown in Fig.\ref{fig:SFConv}, the convolution process is defined as:
\begin{equation}
\bm{I}_{out} = \bm{P} \circledast (\bm{Q} \circledast \bm{I}_{in})
\label{equ:io2},
\end{equation} 
where $\circledast$ denotes the circular convolution process. The reshaping strategy splits a 2d $k \times k$ convolutional filter into two 1d $1 \times k$ convolutional filters to perform the convolution in two steps. Together two 1d convolutions require $\mathcal{O}(kr(c_{in}+c_{out}))$ memory and computation, which is significantly lesser than the $\mathcal{O}(kkc_{in}c_{out})$ cost of unfactorized convolution when $r\ll k \times min(c_{in},c_{out})$. We argue that weights in factorized convolutions are compact and should be effectively utilized. 
Hence, the next section studies how to regularize factorized convolutions to prevent low-rank matrices from large spectral norms.
The strategy is to perform precise inhibition to part of weights by bounding the spectral norm to ensure relatively even contributions.

\subsection{KL Regularizer}


The singular value corresponds to the critical information implicit in the matrix, and the importance is positively related to the magnitude of the singular value \cite{wall2003singular}. Over-large singular values occupy the most informativity of the matrix, thus potentially inducing the bias of image representation. And since the lower pixel-wise variance of inputted medical images is powerless to drive weight updating, it usually occurs over-large singular values of weight matrices or lower weight variance after model training. To bound occurring over-large singular value of weight matrices optimized by medical images with lower pixel-wise variance, we flatten singular values on two low-rank matrices of factorized convolutions by penalizing KL divergence between weight distributions and uniform distribution.
Here, we first perform singular value decomposition on the two matrices $\bm{P}$ and $\bm{Q}$ respectively. The decomposed matrices are expressed as follows:
\begin{equation}
\bm{P} = \bm{U}_{mm} \times \bm{\Sigma}_{mr} \times \bm{V}_{rr}
\label{equ:p},
\end{equation}
\begin{equation}
\bm{Q} = \bm{U}_{rr} \times \bm{\Sigma}_{rn} \times \bm{V}_{nn}
\label{equ:q},
\end{equation}
where $\bm{U}_{mm}$ and $\bm{V}_{rr}$ are unitary matrices obtained by singular value decomposition, consisting of corresponding singular vectors. $\bm{\Sigma}_{mr}$ is a diagonal matrix, and the non-zero values on its diagonal are singular values, arranged in descending order. The meanings of $\bm{U}_{rr}$, $\bm{\Sigma}_{rn}$, and $\bm{V}_{nn}$ are the same. We use $s_{p}$ and $s_{q}$ to represent the vector composed of the singular values in the diagonal matrix $\bm{\Sigma}_{mr}$ and $\bm{\Sigma}_{rn}$. To obtain the distribution of singular values, the elements of the vectors are further converted to $[0,1]$, which are expressed as:
\begin{equation}
s_{p} = \frac{s_{p}}{\bm{F}_{\text{sum}}(s_{p})}
\label{equ:sp},
\end{equation}
\begin{equation}
s_{q} = \frac{s_{q}}{\bm{F}_{\text{sum}}(s_{q})}
\label{equ:sq},
\end{equation}
where $\bm{F}_{\text{sum}}$ represents element-wise summation of the vector. Here, we define a uniform distribution as the all ones vector divided by the length of the corresponding singular value vector. 
$u_{p}$ and $u_{q}$ denote all ones vectors with the same length as $s_{p}$ and $s_{q}$, respectively. 
So the uniform distributions corresponding to $s_{p}$ and $s_{q}$ are expressed as:
\begin{equation}
u_{p} = \frac{u_{p}}{\bm{F}_{\text{len}}(u_{p})}
\label{equ:sp_1},
\end{equation}
\begin{equation}
u_{q} = \frac{u_{q}}{\bm{F}_{\text{len}}(u_{q})}
\label{equ:sq_1},
\end{equation}
where $\bm{F}_{\text{len}}$ denotes the length of the vector. Then, We utilize the KL divergence to measure how different the two distributions are. Since there are multiple convolution kernels in the network, the parameters of these kernels need to be optimized during backpropagation. Therefore, we leverage summation to accumulate the KL divergence between the singular value vectors and the uniform distribution for all kernels in the network. Suppose there are $c$ convolution kernels in the network, the KL regular term is expressed as follows:
\begin{equation}
\mathcal{L}_{\text{KL}} = \sum_{i=1}^{c}\bm{F}_{\text{KL}}(s^i_{p}, u^i_{p})+\bm{F}_{\text{KL}}(s^i_{q}, u^i_{q})  \quad (i=1,...,c)
\label{equ:sp_2},
\end{equation}
where $\bm{F}_{\text{KL}}$ is the KL divergence. To combat the over-large singular values of the weight matrix optimized by medical images, we further penalize the KL regularization term to make the weight distribution tend to be uniform. Therefore, the final optimization goal of the network is expressed as:
\begin{equation}
\mathcal{L} = \mathcal{L}_{\text{loss}} + \lambda \mathcal{L}_{\text{KL}}
\label{equ:sp_3},
\end{equation}
where $\mathcal{L}_{\text{loss}}$ is the loss term and $\lambda$ is the factor used to balance the KL regularization term $\mathcal{L}_{\text{KL}}$. The setting of $\lambda$ is determined empirically. In this way, the weight distribution is optimized during the backpropagation process, which improves the driving ability of the medical image to the model.

\section{Experiments}
\label{sec:exp}
\subsection{Experiments Settings}

\textbf{Datasets.} The classification task is performed on Indian Diabetic Retinopathy Image Dataset (IDRiD)~\cite{porwal2020idrid}. For the segmentation task, we use subset ROSE-1 in ROSE~\cite{ma2020rose}.

\textit{IDRiD.} 
This dataset contains a training set of $413$ fundus images and a test set of $103$ images. 
Each example is a $2848\times4288$ image and is labeled by the risk of macular edema, which has 3 grade in total. We resize those images into $324\times324$ pixels before training. The data preprocessing reference work~\cite{zeng2022factorized}.

\textit{ROSE-1.} 
This dataset contains OCTA images of superficial vascular complexes (SVC) and deep vascular complexes (DVC). We perform the vessel segmentation on SVC, which has a training set of 30 images and a testing set of 9 images. The resolution is $304\times304$ pixels.
\\
\textbf{Optimizer.} 
All experiments are performed on NVIDIA TITAN RTX GPUs. All models are trained with Adam optimizer. For classification, the learning rate is $0.005$ and weight decay is $0.00001$. For segmentation, the learning rate is $0.01$ and weight decay is $0.00001$. The gamma of the learning rate scheduler is $1$ and the step size is $10$ in all experiments.
\\
\textbf{Backbones.} 
We use the classification network Res-Net~\cite{he2016deep} and the segmentation network U-Net~\cite{ronneberger2015u} as the backbone of the classification and segmentation tasks, respectively.
\\
\textbf{Comparable Methods.} The corresponding convolutional layers in the network are replaced by different convolutions in comparative experiments. We compare our SFConv with standard convolution (Conv), the factorized convolution (FConv) applying spectral initialization and frobenius decay~\cite{khodak2021initialization}, the factorized convolution with spectral normalization (FConvSN)~\cite{zeng2022factorized}, and depthwise separable convolution (DPConv)~\cite{chollet2017xception}. 
In addition, we compare SFConv without KL regularizer for ablation study to observe its effect.

\subsection{Evaluation of Fundus Classification}
In this task, we perform the classification of the risk of macular edema. 
The cross-entropy loss function is used to guide the model training. The batch size is set to $32$ and the epoch is $100$. The rank scale of FConv is $0.5$. The latent factor of FConvSN and our SFConv is set to $10$. The balance coefficient $\lambda$ of $\mathcal{L}_{\text{KL}}$ is set to $5$. Final result is shown in Table~\ref{Result:cls}.

\textit{Complexity.}
As shown in Table~\ref{Result:cls}, we see that SFConv has the lowest parameter quantity. This demonstrates that SFConv has powerful model compressing ability by factorizing convolutional layers. SFConv utilizes a two-step convolution to replace the low-rank decomposition, leading to higher FLOPs. 
The FPS of SFConv reaches $70\%$ of the standard convolution, demonstrating that the two-step convolution strategy adopted by our SFConv avoids the decline of inference speed caused by matrix multiplication calculation.

\textit{Performance.}
SFConv matches standard convolutions in performance with lower model complexity.  
Our method prevents large singular values and promotes uniformity of the weight matrix to combat the lower pixel variance of medical images. It further enhances the expressiveness of factorized convolutions in MIP models.

\begin{table}[t]
\caption{The classification results of different methods.\textcolor{white}{g}}
\label{Result:cls}
\begin{tabular}{|
>{\columncolor[HTML]{FFFFFF}}c |
>{\columncolor[HTML]{FFFFFF}}c |
>{\columncolor[HTML]{FFFFFF}}c |
>{\columncolor[HTML]{FFFFFF}}c |
>{\columncolor[HTML]{FFFFFF}}c |}
\toprule
\hline
\cellcolor[HTML]{FFFFFF}\textbf{Methods} & \textbf{\#.Params} & \textbf{FPS} & \textbf{FLOPs} & \textbf{ACC}   \\ \hline
Conv                                        & 11.22M & \textbf{222.23} & 43.03G         & 0.8058 \\
FConv~\cite{khodak2021initialization}       & 2.78M  & 38.12           & {\ul 3.07G}          & 0.7670 \\
FConvSN~\cite{zeng2022factorized}           & {\ul0.40M}  & 84.67           & \textbf{2.78G} & 0.7961 \\
DPConv~\cite{chollet2017xception}           & 1.48M  & {\ul204.27}          & 105.11G        & {\ul 0.8155} \\ \hline
SFConv w/o KL                               & 0.27M  & 159.72          & 20.08G         & 0.7379 \\
\textbf{SFConv(ours)}                       & \textbf{0.27M}     & 158.15       & 20.08G   & \textbf{0.8252} \\ 
\hline
\bottomrule
\end{tabular}
\end{table}


\subsection{Evaluation of OCTA Segmentation}

In this task, we perform vessel segmentation on OCTA images. We utilize dice loss and dice coefficient for model training and performance evaluation. The batch size is set to 16 and the max epoch is 50. The balance coefficient $\lambda$ of $\mathcal{L}_{\text{KL}}$ is set to 10, and all other parameter settings remain the same as classification. Final result is displayed in Table~\ref{Result:seg}.

\begin{figure}[t]
\centering
\includegraphics[width=1.0\linewidth]{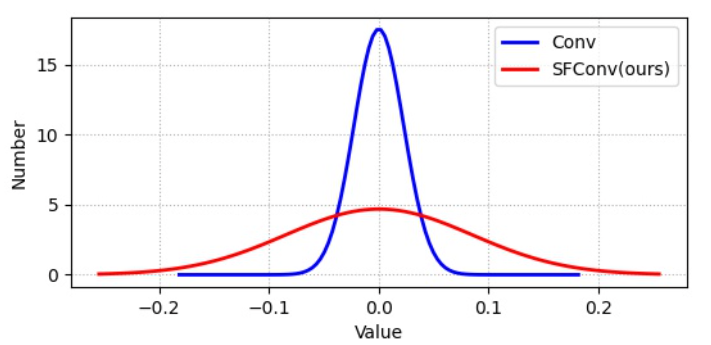} 
\caption{Parameter distributions of the trained U-Net.}
\label{fig:Param}
\end{figure}

The complexity of SFConv in blood vessel segmentation matches that of classification, showcasing its suitability for network lightweighting across various tasks.
In terms of performance, SFConv slightly trails DPConv but significantly streamlines the model. 
Furthermore, SFConv's performance notably enhances with the KL regularizer, highlighting the effectiveness of our proposed KL regularizer in boosting the driving ability of MIP models.
We compare the parameter distribution of the trained U-Net driven by Conv and SFConv. As shown in Fig.~\ref{fig:Param}, the parameter distribution of the network driven by SFConv has a larger variance. This shows that after equalizing the singular value distribution with the KL regularizer, the parameter distribution is wider. The information of the parameter matrix is no longer occupied by an excessively large singular value, so the performance of SFConv is significantly improved.


The above experimental analysis shows that our proposed SFConv with KL regularizer effectively reduces the model complexity. To a certain extent, it solves the problem of weak driving ability of the model caused by the skewed pixel value distribution and low variance of medical image.

\begin{table}[t]
\caption{The segmentation results of different methods.}
\label{Result:seg}
\begin{tabular}{|
>{\columncolor[HTML]{FFFFFF}}c |
>{\columncolor[HTML]{FFFFFF}}c |
>{\columncolor[HTML]{FFFFFF}}c |
>{\columncolor[HTML]{FFFFFF}}c |
>{\columncolor[HTML]{FFFFFF}}c |}
\toprule
\hline
\cellcolor[HTML]{FFFFFF}\textbf{Methods} & \textbf{\#.Params} & \textbf{FPS} & \textbf{FLOPs} & \textbf{Dice} \\ \hline
Conv                              & 17.29M         & \textbf{232.10} & 57.99G         & 0.7476                \\
FConv~\cite{khodak2021initialization}   & 3.60M  & 73.30           & 0.21G          & 0.6770          \\
FConvSN~\cite{zeng2022factorized}           & {\ul 0.56M}  & 76.22           & \textbf{0.21G} & 0.7503          \\
DPConv~\cite{chollet2017xception}           & 2.03M  & {\ul 166.94}          & 7.22G          & \textbf{0.7738} \\ \hline
SFConv w/o KL                               & 0.32M  & 155.65          & 3.08G          & 0.7327          \\
\textbf{SFConv(ours)}      & \textbf{0.32M}     & 156.57       & {\ul 3.08G}    & {\ul 0.7652}  \\ 
\hline
\bottomrule
\end{tabular}
\end{table}



\section{Conclusions}
\label{sec:con}


To alleviate the common problems of slow inference and performance degradation in existing factorized convolutions, we propose a novel factorized convolution with a singular value equalization generalizer (SFConv) for efficient and high-quality model deployment. 
Extensive experiments on fundus images and OCTA datasets demonstrate that our SFConv is highly competitive among vanilla convolutions and numerous lightweight convolutions.

\vfill\pagebreak

\bibliographystyle{IEEEbib}
\bibliography{strings,refs}

\begin{thebibliography}{10}

\bibitem{han2015learning}
Song Han, Jeff Pool, John Tran, and William Dally,
\newblock ``Learning both weights and connections for efficient neural
  network,''
\newblock {\em Advances in neural information processing systems}, vol. 28,
  2015.

\bibitem{hinton2015distilling}
Geoffrey Hinton, Oriol Vinyals, Jeff Dean, et~al.,
\newblock ``Distilling the knowledge in a neural network,''
\newblock {\em arXiv preprint arXiv:1503.02531}, vol. 2, no. 7, 2015.

\bibitem{zhang2015accelerating}
Xiangyu Zhang, Jianhua Zou, Kaiming He, and Jian Sun,
\newblock ``Accelerating very deep convolutional networks for classification
  and detection,''
\newblock {\em IEEE transactions on pattern analysis and machine intelligence},
  vol. 38, no. 10, pp. 1943--1955, 2015.

\bibitem{wang2017factorized}
Min Wang, Baoyuan Liu, and Hassan Foroosh,
\newblock ``Factorized convolutional neural networks,''
\newblock in {\em Proceedings of the IEEE International Conference on Computer
  Vision Workshops}, 2017, pp. 545--553.

\bibitem{li2019compressing}
Tuanhui Li, Baoyuan Wu, Yujiu Yang, Yanbo Fan, Yong Zhang, and Wei Liu,
\newblock ``Compressing convolutional neural networks via factorized
  convolutional filters,''
\newblock in {\em Proceedings of the IEEE/CVF Conference on Computer Vision and
  Pattern Recognition}, 2019, pp. 3977--3986.

\bibitem{denton2014exploiting}
Emily~L Denton, Wojciech Zaremba, Joan Bruna, Yann LeCun, and Rob Fergus,
\newblock ``Exploiting linear structure within convolutional networks for
  efficient evaluation,''
\newblock {\em Advances in neural information processing systems}, vol. 27,
  2014.

\bibitem{khodak2021initialization}
Mikhail Khodak, Neil Tenenholtz, Lester Mackey, and Nicolo Fusi,
\newblock ``Initialization and regularization of factorized neural layers,''
\newblock {\em arXiv preprint arXiv:2105.01029}, 2021.

\bibitem{zeng2022factorized}
Ming Zeng, Na~Zeng, Jiansheng Fang, and Jiang Liu,
\newblock ``Factorized convolution with spectral normalization for fundus
  screening,''
\newblock in {\em 2022 IEEE 19th International Symposium on Biomedical Imaging
  (ISBI)}. IEEE, 2022, pp. 1--5.

\bibitem{everingham2015pascal}
Mark Everingham, SM~Ali Eslami, Luc Van~Gool, Christopher~KI Williams, John
  Winn, and Andrew Zisserman,
\newblock ``The pascal visual object classes challenge: A retrospective,''
\newblock {\em International journal of computer vision}, vol. 111, no. 1, pp.
  98--136, 2015.

\bibitem{nguyen2020vindr}
Ha~Q Nguyen, Khanh Lam, Linh~T Le, Hieu~H Pham, Dat~Q Tran, Dung~B Nguyen,
  Dung~D Le, Chi~M Pham, Hang~TT Tong, Diep~H Dinh, et~al.,
\newblock ``Vindr-cxr: An open dataset of chest x-rays with radiologist's
  annotations,''
\newblock {\em arXiv preprint arXiv:2012.15029}, 2020.

\bibitem{doane2011measuring}
David~P Doane and Lori~E Seward,
\newblock ``Measuring skewness: a forgotten statistic?,''
\newblock {\em Journal of statistics education}, vol. 19, no. 2, 2011.

\bibitem{tai2015convolutional}
Cheng Tai, Tong Xiao, Yi~Zhang, Xiaogang Wang, et~al.,
\newblock ``Convolutional neural networks with low-rank regularization,''
\newblock {\em arXiv preprint arXiv:1511.06067}, 2015.

\bibitem{wall2003singular}
Michael~E Wall, Andreas Rechtsteiner, and Luis~M Rocha,
\newblock ``Singular value decomposition and principal component analysis,''
\newblock in {\em A practical approach to microarray data analysis}, pp.
  91--109. Springer, 2003.

\bibitem{porwal2020idrid}
Prasanna Porwal, Samiksha Pachade, Manesh Kokare, Girish Deshmukh, Jaemin Son,
  Woong Bae, Lihong Liu, Jianzong Wang, Xinhui Liu, Liangxin Gao, et~al.,
\newblock ``Idrid: Diabetic retinopathy--segmentation and grading challenge,''
\newblock {\em Medical image analysis}, vol. 59, pp. 101561, 2020.

\bibitem{ma2020rose}
Yuhui Ma, Huaying Hao, Jianyang Xie, Huazhu Fu, Jiong Zhang, Jianlong Yang,
  Zhen Wang, Jiang Liu, Yalin Zheng, and Yitian Zhao,
\newblock ``Rose: a retinal oct-angiography vessel segmentation dataset and new
  model,''
\newblock {\em IEEE transactions on medical imaging}, vol. 40, no. 3, pp.
  928--939, 2020.

\bibitem{he2016deep}
Kaiming He, Xiangyu Zhang, Shaoqing Ren, and Jian Sun,
\newblock ``Deep residual learning for image recognition,''
\newblock in {\em Proceedings of the IEEE conference on computer vision and
  pattern recognition}, 2016, pp. 770--778.

\bibitem{ronneberger2015u}
Olaf Ronneberger, Philipp Fischer, and Thomas Brox,
\newblock ``U-net: Convolutional networks for biomedical image segmentation,''
\newblock in {\em International Conference on Medical image computing and
  computer-assisted intervention}. Springer, 2015, pp. 234--241.

\bibitem{chollet2017xception}
Fran{\c{c}}ois Chollet,
\newblock ``Xception: Deep learning with depthwise separable convolutions,''
\newblock in {\em Proceedings of the IEEE conference on computer vision and
  pattern recognition}, 2017, pp. 1251--1258.

\end{thebibliography}

\end{document}